\documentclass{article}
\usepackage{ijcai25}
\usepackage{times}
\usepackage{soul}
\usepackage{tikz}
\usetikzlibrary{shapes.geometric, arrows, positioning,fit,calc}
\usepackage{url}
\usepackage[hidelinks]{hyperref}
\usepackage[utf8]{inputenc}
\usepackage[small]{caption}
\usepackage{graphicx}
\usepackage{amsthm}

\usepackage{amsmath,amssymb,amsfonts}
\usepackage{booktabs}
\usepackage{subcaption}
\usepackage{algorithm}
\usepackage{algorithmic}

\usepackage{multirow}
\usepackage{tabularx}
\usepackage{enumitem}
\usepackage{bbm}
\usepackage{balance}
\usepackage[flushleft]{threeparttable}

\usepackage{color, colortbl}
\definecolor{Gray}{gray}{0.9}
\usepackage{setspace}

\title{A Survey on Diffusion Models for Anomaly Detection}

\author{
Jing Liu$^{1,2}$\and
Zhenchao Ma$^1$\and
Zepu Wang$^3$\and
Chenxuanyin Zou$^1$\and
Jiayang Ren$^1$\and\\
Zehua Wang$^1$\and
Liang Song$^2$\and 
Bo Hu$^2$\and
Yang Liu$^{2,3}$\and
Victor C.M. Leung$^1$
\affiliations
$^1$University of British Columbia, Vancouver, Canada\\
$^2$Fudan University, Shanghai, China\\
$^3$Duke Kunshan University, Suzhou, China\\
\emails
\{jingliu, zhenchaoma, zwang\}@ece.ubc.ca, \{zcxy, rjy12307\}@mail.ubc.ca,\\zw163@duke.edu, \{songl, bohu, yang\_liu20\}@fudan.edu.cn, vleung@ieee.org
}

\begin{document}
\maketitle
\begin{abstract}

Diffusion models (DMs) have emerged as a powerful class of generative AI models, showing remarkable potential in anomaly detection (AD) tasks across various domains, such as cybersecurity, fraud detection, healthcare, and manufacturing. The intersection of these two fields, termed diffusion models for anomaly detection (DMAD), offers promising solutions for identifying deviations in increasingly complex and high-dimensional data. In this survey, we review recent advances in DMAD research. We begin by presenting the fundamental concepts of AD and DMs, followed by a comprehensive analysis of classic DM architectures including DDPMs, DDIMs, and Score SDEs. We further categorize existing DMAD methods into reconstruction-based, density-based, and hybrid approaches, providing detailed examinations of their methodological innovations. We also explore the diverse tasks across different data modalities, encompassing image, time series, video, and multimodal data analysis. Furthermore, we discuss critical challenges and emerging research directions, including computational efficiency, model interpretability, robustness enhancement, edge-cloud collaboration, and integration with large language models. The collection of DMAD research papers and resources is available at \texttt{\url{https://github.com/fdjingliu/DMAD}}.

\end{abstract}
\section{Introduction}\label{sec1}

The ever-increasing volume, velocity, and variety of data present both opportunities and challenges for anomaly detection (AD).  In cybersecurity, real-time threat detection is crucial given the constant influx of network traffic and logs \cite{jin2024survey}.  Similarly, robust fraud detection systems are essential for financial institutions processing massive transaction datasets \cite{wang2023drift}.  Healthcare and manufacturing also rely heavily on data for early disease diagnosis and predictive maintenance, respectively \cite{zhang2023diffusionad}. Such applications highlight the growing need for automated AD methods to efficiently identify outliers in complex datasets. However, traditional techniques, often based on statistical methods or rule-based systems, struggle with the scale and complexity of modern data \cite{katsuoka2024statistical}.  Specifically, these methods often require extensive manual feature engineering and struggle to adapt to evolving data distributions.  Additionally, the sheer data volume can overwhelm traditional methods, rendering them computationally infeasible for real-time applications. Consequently, more sophisticated and scalable AD approaches are needed. For example, weakly supervised approaches leverage limited labeled data to improve performance, driving innovation toward scalable deep learning architectures that address these fundamental challenges \cite{pang2021deepa}.

Diffusion models (DMs) have emerged as a powerful class of generative models, synthesizing samples across diverse data modalities \cite{wang2023diffusion}. Unlike generative adversarial networks (GANs) and variational autoencoders (VAEs), which may exhibit training instability or produce less detailed samples \cite{dao2024highquality}, DMs generate sharper, more realistic samples through a gradual denoising process \cite{chen2024diffilter}. Consequently, DMs excel in capturing a broader range of data distributions compared to GANs \cite{wyatt2022anoddpm}, making them particularly suitable for anomaly detection, where their enhanced mode coverage facilitates precise modeling of normal patterns and subsequent identification of distributional outliers \cite{li2024selfsupervised}.

Recent growth in complex and multi-dimensional data has created new challenges for anomaly detection methods. In this context, DMs have emerged as a promising solution due to their inherent connection to density estimation \cite{lelan2021perfect}. DMs learn the probability distributions of normal data through iterative denoising. Leveraging their exceptional generative abilities, DMs accurately reconstruct normal patterns while capturing underlying manifold structures, enabling anomaly detection through reconstruction error analysis and probability density estimation \cite{hu2023anodode,salimans2021progressive}.  For example, AnoDDPM \cite{wyatt2022anoddpm} uses diffusion models for anomaly detection via reconstruction error, and ODD \cite{wang2023odd} enhances this by leveraging advanced generative modeling for high-dimensional data.  Additionally, the learned score function, representing the gradient of the log-probability density, can be directly used as an anomaly score, as demonstrated in diffusion time estimation \cite{livernoche2023diffusion}.

\paragraph{Motivations.} While existing surveys have separately reviewed anomaly detection~\cite{jin2024survey,pang2021deepa} and diffusion models~\cite{yang2024survey,luo2023comprehensive}, they either focus on traditional deep learning approaches without considering the emerging role of diffusion models, or treat anomaly detection merely as one of many DM applications without in-depth analysis. Recent breakthroughs in diffusion models for anomaly detection (DMAD) demonstrate remarkable potential to revolutionize the field through their unique capabilities in modeling complex data distributions and generating high-quality samples. Despite these advances, there is no existing work has systematically reviewed how DMs enhance anomaly detection or analyzed their emerging role at the intersection of both domains. Our work addresses this gap by providing a timely overview of DMAD, analyzing its theoretical foundations, current achievements, and promising future directions that could fundamentally reshape the landscape of anomaly detection research.

\begin{figure}[t!]
    \centering
	\includegraphics[width=0.48\textwidth]{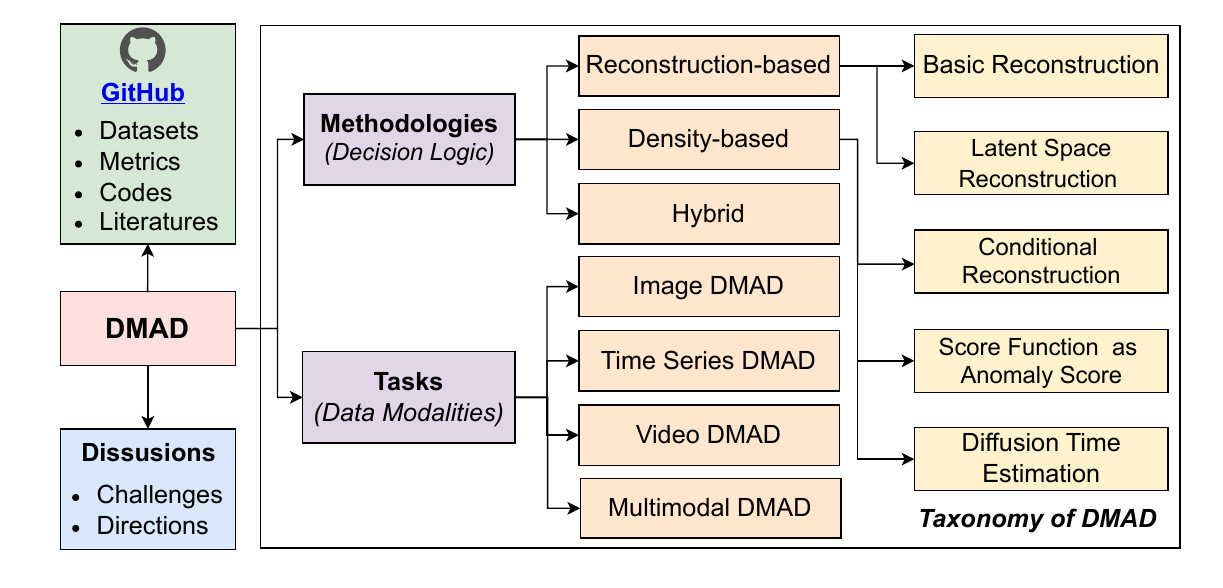}
    \caption{Taxonomy of diffusion models for anomaly detection.}
    \label{fig:1}
\end{figure}

\paragraph{Contributions.} The main contributions of this survey are four-fold: \textbf{(1)} \textit{A systematic taxonomy}. We present the first comprehensive taxonomy of DMAD (as shown in Fig.~\ref{fig:1}), categorizing existing methodologies into reconstruction-based, density-based, and hybrid methods, providing a structured framework for understanding this emerging field. \textbf{(2)} \textit{A comprehensive review}. Based on the proposed taxonomy, we systematically analyze DMAD across different tasks, including image anomaly detection (IAD), time series anomaly detection (TSAD), video anomaly detection (VAD), and multimodal anomaly detection (MAD), discussing their approaches, strengths, and limitations. \textbf{(3)} \textit{Technical challenges and future directions}. We identify and discuss critical challenges in DMAD, including computational costs, interpretability, robustness against adversarial attacks, complex data distributions, edge-cloud collaboration, and integration with large language models (LLMs), while suggesting potential solutions and future research directions. \textbf{(4)} \textit{Resources and benchmarks}. We provide a thorough collection of representative methods, implementations, datasets, and evaluation metrics for different tasks, serving as a valuable resource for researchers and practitioners in this field.

\section{Preliminaries}\label{sec2}
This section begins by providing the fundamental concepts of anomaly detection. Following this, we introduce the various types of anomalies, including point, contextual, and collective anomalies. Next, we discuss the basic principles of DMs and their relevance to anomaly detection. We then explore score-based models and score matching techniques. Finally, we present prominent variants of DMs used in anomaly detection, highlighting their advantages and disadvantages.

\subsection{Anomaly Detection}\label{sec2.1}
Anomaly detection aims to identify instances deviating significantly from the expected data distribution \cite{wang2023hidden}. Given a dataset $\mathcal{D} = \{x_1, x_2, ..., x_n\}$, where $x_i \in \mathbb{R}^d$, the goal is to identify the anomalous subset $\mathcal{A} \subset \mathcal{D}$.  Anomalous behavior is characterized by low probability density under the typically unknown distribution $p(x)$.  Several anomaly types exist, including point anomalies \cite{hu2023anodode} that are individual deviant instances; contextual anomalies \cite{sui2024anomaly} that are anomalous within a specific context; and collective anomalies that are groups of instances anomalous only when considered together. Common evaluation metrics include AUC \cite{xu2023unsupervised}, measuring the classifier's ability to distinguish between normal and anomalous instances.  Additionally, precision, recall, and F1-score assess the trade-off between correct identification and minimizing false alarms~\cite{wang2023drift,xiao2023imputationbased,zuo2024unsupervised}. The choice of metric depends on the application and the relative importance of minimizing false positives and negatives \cite{pang2021deepa}.\subsection{Diffusion Models}\label{sec2.2}

DMs constitute a class of latent variable models defined by forward and reverse Markov processes to corrupt and denoise data, with training based on the variational lower bound of the negative log-likelihood.

\paragraph{Forward Process.} Forward process is known as the diffusion process, gradually adds noise to data $x_0$ over $T$ timesteps, transforming its distribution into an isotropic Gaussian~\cite{wyatt2022anoddpm}. Specifically, given a variance schedule $\{\beta_t\}_{t=1}^T$, where $0 < \beta_1 < \dots < \beta_T < 1$, the process generates sequence of latent variables $\{x_1, \dots, x_T\}$ via a Markov chain with the Gaussian transition kernel:
\begin{equation}
 q(x_t|x_{t-1}) = \mathcal{N}(x_t; \sqrt{1 - \beta_t} x_{t-1}, \beta_t \mathbf{I}).
\end{equation}
Here, $\mathcal{N}(x; \mu, \Sigma)$ denotes a Gaussian distribution with mean $\mu$ and covariance $\Sigma$, and $\mathbf{I}$ is the identity matrix. The parameter $\beta_t$ controls the noise added at each step.  As $t$ increases, $x_T$ approaches Gaussian noise. Due to the Markov property~\cite{li2023fast}, the joint distribution is:
\begin{equation}
 q(x_{1:T}|x_0) = \prod_{t=1}^T q(x_t|x_{t-1}).
\end{equation}
Consequently, any $x_t$ can be directly sampled from $x_0$:
\begin{equation}
 q(x_t|x_0) = \mathcal{N}(x_t; \sqrt{\bar{\alpha}_t} x_0, (1 - \bar{\alpha}_t) \mathbf{I}),
\end{equation}
where $\alpha_t = 1 - \beta_t$ and $\bar{\alpha}_t = \prod_{i=1}^t \alpha_i$, which provides computational benefits.  Related work explores this process through various perspectives, including multiplicative transitions, uniformization techniques for convergence analysis, Gaussian process covariance transformations, tensor network modeling, first-passage time statistics, and infinite-dimensional diffusion processes \cite{kumar2023selfsupervised}.

\paragraph{Reverse Process.} Reverse process aims to learn the conditional probability distribution $p_\theta(x_{t-1}|x_t)$ to recover the original data distribution by iteratively removing noise.  It is typically modeled as a Gaussian distribution:
\begin{equation}
 p_\theta(x_{t-1}|x_t) = \mathcal{N}(x_{t-1}; \mu_\theta(x_t, t), \Sigma_\theta(x_t, t)),
\end{equation}
where $\mu_\theta$ and $\Sigma_\theta$ are the predicted mean and covariance, respectively, parameterized by a carefully designed neural network (e.g., U-Net \cite{yan2024hybrid} or Transformer \cite{cao2024timedit}) and conditioned on the noisy input $x_t$ at timestep $t$.  While the covariance $\Sigma_\theta(x_t, t)$ can be learned, a common simplification is to use a fixed, time-dependent variance \cite{salimans2021progressive}. The neural network learns to effectively predict the mean $\mu_\theta(x_t, t)$, which guides denoising. Several parameterization strategies exist, including predicting the mean of $q(x_{t-1}|x_t, x_0)$, added noise $\epsilon_t$, or original data $x_0$. Starting from Gaussian noise $x_T$, the reverse process iteratively applies $p_\theta(x_{t-1}|x_t)$ until $x_0$ is reached.

\paragraph{Training Objective.} DMs are trained by maximizing the likelihood of observed data.  However, due to the complex latent variable formulation, direct optimization is computationally intractable. Consequently, training relies on maximizing a variational lower bound (VLB) of the data log-likelihood, derived using variational inference \cite{bercea2023mask} as:

\begin{align} 
	\vspace{-5px}
	\begin{aligned}
		\centering
		\mathcal L_{\rm vlb} = & \mathbb{E}_q[D_{\rm KL}(q(x_T|x_0)||p_\theta(x_T))]-\log p_\theta(x_0|x_1)  \\
		+ & \mathbb{E}_q[\sum_{t=2}^T D_{\rm KL}(q(x_{t-1}|x_{t}, x_0)||p_\theta( x_{t-1}|x_t))],
	\end{aligned}
	\vspace{-5px}
\end{align}
where $q$ and $p_\theta$ denote the forward and learned reverse process distributions, respectively.  Here, $D_{KL}$ is the standard Kullback-Leibler divergence, $x_0$ represents the original data, and $x_t$ is the noisy sample at timestep $t$. A simplified objective function minimizes the $L_2$ distance between true and predicted noise \cite{han2024neural}, can be expressed as:
\begin{equation} 
    \mathcal L_{\rm simple}=\sum_{t=1}^T\mathbb{E}_q \big[||\epsilon_t-\epsilon_{\theta}(x_t, t)||^2 \big],
\end{equation}
where $\epsilon_t$ and $\epsilon_{\theta}(x_t, t)$ denote the true and predicted noise, respectively.  Such simplified objective exhibits computational efficiency and effectiveness \cite{livernoche2023diffusion}.  Additionally, weighting loss terms at different timesteps can improve performance \cite{choi2022perception}.  From another perspective, DMs can be viewed as distribution estimators with theoretical guarantees, and the training objective can be based on score matching, where the model estimates the score function \cite{zhang2024analyzing}.  A unified framework for discrete and continuous-time DMs offers further insights into objective optimization \cite{lee2023codi}.

\subsection{Score-based Models and Score Matching}\label{sec2.3}
\paragraph{Score Matching for Learning Score Functions.} Score-based generative models learn the score function $\nabla_x \log p(x)$, representing the gradient of the log-probability density.  By encoding distributional characteristics directly through these gradients, score functions enable powerful generative capabilities without explicit density estimation. Score matching offers a practical way to learn this score function without needing $p(x)$ explicitly.  Instead, it minimizes the Fisher divergence between the learned and true score functions \cite{han2024neural}. Specifically, the score matching objective is $L_{SM}(\theta) = \frac{1}{2} \mathbb{E}_{p(x)} \left[ \| \nabla_x \log p_\theta(x) - \nabla_x \log p(x) \|^2 \right]$.

\paragraph{Denoising Score Matching.}  However, computing the true score is often intractable. Consequently, denoising score matching perturbs data with Gaussian noise and learning the score of this perturbed distribution.  The objective becomes $L_{DSM}(\theta) = \frac{1}{2} \mathbb{E}_{q_\sigma(x)} \left[ \| \nabla_x \log p_\theta(x) - \nabla_x \log q_\sigma(x) \|^2 \right]$, where $q_\sigma(x)$ is the perturbed distribution.  Learning scores at various noise levels allows DMs to capture the data distribution at different scales \cite{wang2023ensemble}.

\paragraph{Advanced Techniques.}  Recent research highlights the inherent linear structure within score-based models, particularly at higher noise levels, potentially accelerating sampling \cite{wang2023hidden}.  Prioritizing selected specific noise levels during training, based on perceptual relevance, can further improve performance \cite{choi2022perception}.  In addition, gradient guidance, incorporating an external objective function's gradient during sampling, effectively adapts pre-trained models to specific tasks \cite{tur2023unsupervised}.

\subsection{Variants of Diffusion Models}\label{sec2.4}
In this section, we discuss prominent diffusion model variants for anomaly detection, including denoising diffusion probabilistic models (DDPMs), denoising diffusion implicit models (DDIMs), and score-based generative models with stochastic differential equations (Score SDEs), along with their advantages and disadvantages.

\paragraph{DDPMs.} DDPMs~\cite{bercea2023mask} progressively introduce controlled Gaussian noise to the data across multiple time steps using a Markov chain. A trained neural network then reverses this process, enabling the model to effectively learn complex distributions, which makes it particularly well-suited for anomaly detection \cite{wyatt2022anoddpm}. However, generating samples can be computationally expensive.

\paragraph{DDIMs.} DDIMs~\cite{xu2023unsupervised} extend DDPMs with a non-Markovian sampling approach, enabling faster generation by skipping steps.  Specifically, they modify the reverse process, and by setting a specific parameter to zero, DDIMs allow deterministic sampling, beneficial for consistent reconstructions in anomaly detection \cite{tebbe2024dynamic}.

\paragraph{Score SDEs.} Score SDEs~\cite{deveney2023closing} model diffusion as a continuous-time SDE, where the score function, representing the gradient of the log-probability density, can be effectively used for anomaly scoring \cite{li2024selfsupervised}. While offering flexibility for complex distributions \cite{livernoche2023diffusion}, training and parameterization choices can be particularly  challenging \cite{graham2023unsupervised}.

\section{Methodologies}\label{sec3}
Recent research efforts on DMAD can be broadly categorized into three primary methodologies: reconstruction-based, density-based, and hybrid approaches.
\subsection{Reconstruction-based Anomaly Detection}\label{sec3.1}
\begin{figure}[t!]
    \centering
	\includegraphics[width=0.48\textwidth]{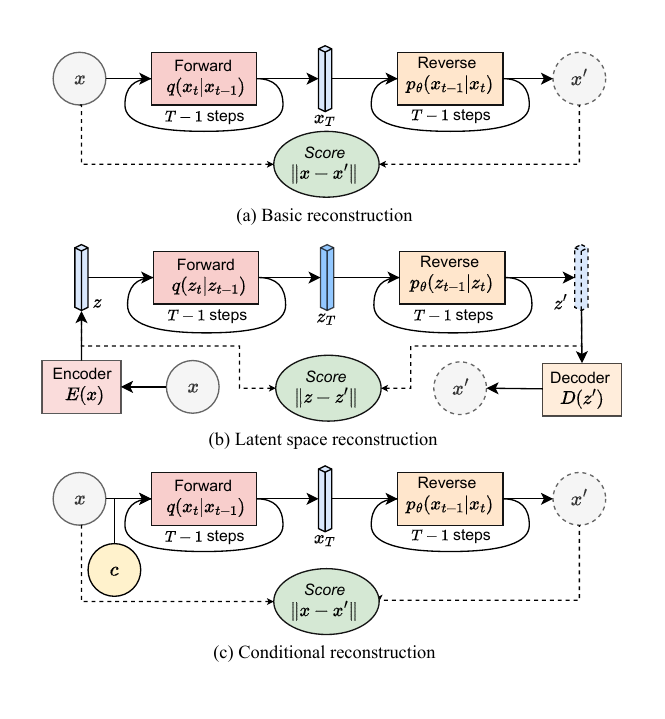}
	\caption{Pipeline illustration of reconstruction-based AD methods: (a) basic, (b) latent space, and (c) conditional reconstruction.}
    \label{fig:2}
\end{figure}

Reconstruction-based AD (RAD) leverages DMs to identify deviations from learned data distributions.  Within this paradigm, we consider three reconstruction strategies, as illustrated in Fig.~\ref{fig:2}: 1) basic reconstruction, where reconstruction error serves as the anomaly score; 2) latent space reconstruction, incorporating dimensionality reduction for efficiency with high-dimensional data; 3) conditional reconstruction, leveraging auxiliary information like class labels for potentially more refined anomaly detection.

\paragraph{Basic Reconstruction.} RAD hinges on the fundamental principle that DMs trained on normal data reconstruct normal instances effectively, but struggle with anomalous ones \cite{hu2023anodode}.  Specifically, after training on normal samples, the model reconstructs a test sample by reversing the diffusion process. The difference between the input and its reconstruction constitutes the anomaly score \cite{zhang2023diffusionad}. A higher score indicates a greater likelihood of an anomaly.  For instance, GLAD \cite{yao2025glad} employs global and local reconstruction perspectives.  In contrast, AutoDDPM \cite{bercea2023mask} focuses on improving normal region reconstruction.  Alternatively, some approaches use a learned similarity network for comparison, yielding a semantic distance as the anomaly score \cite{wang2023odd}.

\paragraph{Latent Space Reconstruction.} RAD leverages latent space representations for enhanced efficiency and effective handling of high-dimensional data, such as images or videos. Specifically, projecting input data into a lower-dimensional latent space significantly reduces the computational burden associated with reconstruction. Autoencoders (AEs) are commonly employed for this dimensionality reduction, which learns a compressed representation by encoding data into a latent space and then decoding it back to the original space.  Integrating AEs with DMs offers a promising approach. For example, DMs trained on AEs-learned latent representations can capture the normal data distribution in the compressed space \cite{hu2023anodode}.  In addition, NGLS-Diff \cite{han2024diffusion} is a novel approach that utilizes DMs within a normal gathering latent space to enhance anomaly detection capabilities for time series data.  The latent space reconstruction error then serves as an anomaly score, with deviations indicating potential anomalies \cite{wang2023diffusion}. Shared latent spaces across multiple AEs can separate treatment and context.  Similarly, supervised dimensionality reduction can optimize the latent space for specific classification tasks, and employing Brownian motion within Diffusion VAEs captures dataset topology \cite{lelan2021perfect}.

\paragraph{Conditional Reconstruction.} Conditional information, such as class labels, masks, or text descriptions, guides the diffusion model's reconstruction toward expected normal outputs.  For example, FDAE \cite{zhu2024flowguided} utilizes foreground objects and motion information as inputs to train a conditional diffusion autoencoder, enhancing the detection of anomalous regions in an unsupervised manner.  Dual conditioning \cite{zhan2024enhancing} improves multi-class anomaly detection by ensuring prediction and reconstruction accuracy within the expected category.  A learnable encoder in RecDMs \cite{xu2023unsupervised} extracts semantic representations for conditional denoising, thus guiding recovery and avoiding trivial reconstructions.  Similarly, GLAD \cite{yao2025glad} introduces synthetic anomalies during training to encourage the model to learn complex noise distributions for improved anomaly-free reconstruction.  In addition, \cite{tebbe2024dynamic} explores dynamic step size computation based on initial anomaly predictions for refined reconstruction.  Finally, MDPS \cite{wu2024unsupervised} models normal image reconstruction as multiple diffusion posterior sampling using a masked noisy observation model and a diffusion-based prior, consequently improving reconstruction quality.

\subsection{Density-based Anomaly Detection}\label{sec3.2}

\begin{figure}[t!]
    \centering
	\includegraphics[width=0.48\textwidth]{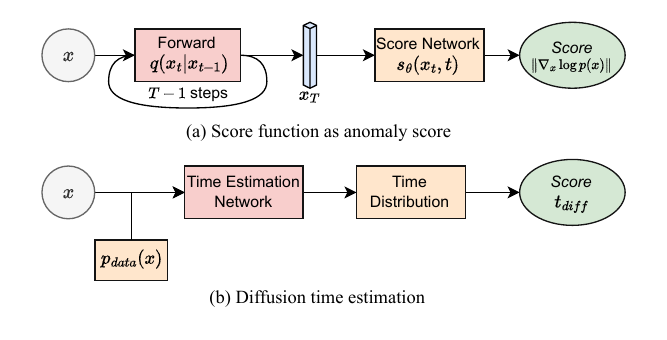}
	\caption{Pipeline illustration of density-based AD methods: (a) score function-based and (b) diffusion time estimation method.}
    \label{fig:3}
\end{figure}

This section considers two primary methods for density-based AD (DAD), as shown in Fig.~\ref{fig:3}: Score function and diffusion time estimation (DTE), illustrated by their use of the learned probability density to identify outliers.

\paragraph{Score Function as Anomaly Score.} Gradient-based log probability density score functions, $\|\nabla_x \log p(x)\|$, accurately and effectively quantify data point likelihoods within learned distributions \cite{zhang2024analyzing}. Within low-density regions, higher score magnitudes reliably indicate potential anomalies by leveraging the score network $s_\theta(x_t,t)$ \cite{livernoche2023diffusion}. Additionally, probability density characterization through score computation enables direct anomaly assessment without requiring explicit reconstruction or comparative analysis. Theoretical foundations established by \cite{wang2023hidden} and \cite{han2024neural} strongly validate this methodology, although recent investigations \cite{deveney2023closing} illuminate potential disparities between SDE and ODE formulations, warranting careful consideration in score-based anomaly detection frameworks.

\paragraph{Diffusion Time Estimation.} DTE offers an alternative anomaly detection method leveraging the diffusion process \cite{livernoche2023diffusion}. Instead of reconstructing the input $x$, DTE estimates the diffusion timestep $t_{diff}$ required for $x$ to traverse the data distribution $p_{data}(x)$. The estimation is performed by a time estimation network, which predicts the time distribution. Intuitively, a longer $t_{diff}$ suggests an anomaly due to the $x$'s distance from the learned distribution \cite{luo2023comprehensive}. The estimated diffusion time $t_{diff}$ thus serves as an anomaly score, derived from the mode or mean of its distribution. An analytical form for this density is derivable, and deep neural networks can improve inference efficiency. Consequently, DTE has shown competitive performance, especially in speed, on benchmarks like ADBench, while maintaining or exceeding the accuracy of traditional DDPMs.

\subsection{Hybrid Approaches}\label{sec3.3}

Hybrid approaches integrate DMs with other anomaly detection techniques to improve performance. For example, RAD can be combined with density estimation, where reconstruction error from DMs is integrated with DAD, like DTE \cite{livernoche2023diffusion}. Additionally, dynamic step size computation in the forward process, guided by initial anomaly predictions, improves performance by denoising scaled input without added noise \cite{tebbe2024dynamic}. In another approach, likelihood maps of potential anomalies from DMs are integrated with the original image via joint noised distribution re-sampling, enhancing healthy tissue restoration \cite{bercea2023mask}.  For TSAD, hybrid methods can employ density ratio-based strategies to select normal observations for imputation, combined with denoising diffusion-based imputation to improve missing value generation, especially under anomaly concentration \cite{xiao2023imputationbased}.  Advanced reconstruction techniques predict specific denoising steps by analyzing divergences between image and diffusion model priors, supplemented by synthetic abnormal sample generation during training and spatial-adaptive feature fusion during inference \cite{yao2025glad}.  Integrating DMs with Transformers for multi-class anomaly detection, where diffusion provides high-frequency information for refinement, mitigates blurry reconstruction and "identical shortcuts" \cite{zhan2024enhancing}.  Finally, a Bayesian framework employing masked noisy observation models and diffusion-based normal image priors enables effective difference map computation between normal posterior samples and the test image, improving anomaly detection and localization \cite{wu2024unsupervised}.
\section{Tasks}\label{sec4}
In this section, we consider four primary tasks of DMAD, as summarized in Tables \ref{tab:1}-\ref{tab:4}, which present representative methods with open-source implementations, datasets, evaluation metrics, and corresponding results. For each task, methods are categorized based on their underlying principles and detection strategies, highlighting the unique challenges and solutions developed for different data modalities.
\subsection{Image Anomaly Detection}\label{sec4.1}
\begin{table}[tb!]
    \centering
    \setlength{\tabcolsep}{0.5mm}{
    \resizebox{0.48\textwidth}{!}{
    \begin{tabular}{@{}>{\centering\arraybackslash}p{2.3cm}|c|>{\centering\arraybackslash}p{3cm}|>{\centering\arraybackslash}p{5.3cm}@{}}
        \toprule
        \multicolumn{1}{c|}{\textbf{Method}} & \multicolumn{1}{c|}{\textbf{DS}} & \multicolumn{1}{c|}{\textbf{Datasets}} & \multicolumn{1}{c}{\textbf{Results of metrics on datasets}} \\ \midrule
        FNDM \shortcite{li2023fast} & D & BRATS, ISLES & Dice: 76.21\% (BRATS), 54.44\% (ISLES), VS: 82.28\% (BRATS), etc. \\
        ODD \shortcite{wang2023odd} & D & MNIST, CIFAR-10, etc. & AUROC: 97.8\% (MNIST), 84.7\% (CIFAR-10), etc. \\
        \href{https://mddpm.github.io/}{\textcolor{blue}{mDDPM}} \shortcite{iqbal2024unsupervised} & D & BraTS21, MSLUB & Dice: 53.02\% (BraTS21), 10.71\% (MSLUB), etc. \\
        Dif-fuse \shortcite{fontanella2024diffusion} & D & IST-3, BraTS, WMH & Dice: 0.699 (BraTS21), 0.569 (WMH), etc. \\
        DIC \shortcite{tebbe2024dynamic} & C & VisA, BTAD, MVTec & I-AUROC: 96.0\%, P-AUROC: 97.9\%, PRO 94.1\% on VisA, etc. \\
        \href{https://github.com/samuel_sxy/163.com}{\textcolor{blue}{DRAD}} \shortcite{sheng2024surface} & C & MVTec-AD & I-AUROC: 99.1\% (MVTec-AD), P-AUROC: 97.3\% (MVTec-AD), etc. \\
        \href{https://github.com/KevinBHLin/}{\textcolor{blue}{MDPS}} \shortcite{wu2024unsupervised} & C & MVTec-AD, BTAD & I-AUROC: 98.4\% (MVTec-AD), P-AUROC: 97.0\% (MVTec-AD), etc. \\
        \href{https://github.com/snavalm/disyre}{\textcolor{blue}{DISYRE}} \shortcite{navalmarimont2024ensembled} & C & CamCAN, ATLAS, BraTS & AP: 45\% (ATLAS), 51\% (BraTS-T1), 73\% (BraTS-T2), etc. \\ \bottomrule
    \end{tabular}}}
    \fontsize{8.5pt}{8pt}\selectfont {
    \begin{minipage}{0.48\textwidth}
    \textit{Notes:} \textbf{Method} highlighted in \textcolor{blue}{blue} indicates available open-source code links. \textbf{DS} denotes diffusion space, where 'D' is discrete and 'C' is continuous. Due to space limitations, only partial datasets and results are shown. For more details, please find our \href{https://github.com/fdjingliu/DMAD}{\textcolor{blue}{GitHub}}.
    \end{minipage}
    }
    \caption{Summary of IAD methods with metric results on datasets.}
    \label{tab:1}
\end{table}
IAD represents a key application area for DMs, focusing on identifying deviations from normality in both global patterns affecting entire images and local anomalies confined to specific regions. Many approaches leverage DMs' reconstruction capabilities, with methods calculating anomaly scores from reconstruction errors after reconstructing images from noisy inputs \cite{xu2023unsupervised,tur2023exploring,wang2023ensemble}. Advanced frameworks like GLAD \cite{yao2025glad} employ global and local adaptive approaches, while ODD \cite{wang2023odd} introduces similarity networks for semantic distance measurement. Medical imaging applications have shown particular promise, with FNDM \cite{li2023fast} and mDDPM \cite{iqbal2024unsupervised} demonstrating enhanced efficiency through masked diffusion models, while \cite{fontanella2024diffusion} introduces novel counterfactual generation techniques. Recent research includes AnomalyDiffusion \cite{hu2024anomalydiffusion} for few-shot generation and DISYRE \cite{navalmarimont2024ensembled} for synthetic anomaly generation in medical imaging.. MDPS \cite{wu2024unsupervised} and DRAD \cite{sheng2024surface} further enhance detection through masked sampling and noise embedding, while methods like DIC \cite{tebbe2024dynamic} employ dynamic step size computation to improve localization accuracy.

\subsection{Time Series Anomaly Detection}\label{sec4.2}
\begin{table}[tb!]
	\centering
	\setlength{\tabcolsep}{0.5mm}{
	\resizebox{0.48\textwidth}{!}{
	
		\begin{tabular}{@{}>{\centering\arraybackslash}p{2.5cm}|c|>{\centering\arraybackslash}p{2.95cm}|m{5.3cm}@{}}

		\toprule
		\multicolumn{1}{c|}{\textbf{Method}} & \multicolumn{1}{c|}{\textbf{DS}} & \multicolumn{1}{c|}{\textbf{Datasets}} & \multicolumn{1}{c}{\textbf{Results of metrics on datasets}} \\ \midrule
	\href{https://github.com/yixinliu233/SimpDM}{\textcolor{blue}{SimpDM}} \shortcite{liu2024selfsupervision} & C & Iris, Yacht, etc. & RMSE: 0.059 (Iris), 0.045 (Yacht), 0.102 (Housing), etc. \\
	\href{https://github.com/ChaejeongLee/CoDi}{\textcolor{blue}{CoDi}} \shortcite{lee2023codi} & C \& D & Bank, Heart, etc. & B-F1: 47.26\%, B-AUROC: 81.06\%, M-F1: 62.21\%, etc., on Bank \\
	TimeDiT \shortcite{cao2024timedit} & C & Stocks, Energy, ect. & DS: 11.50\% (Stocks), 17.78\% (Air Quality), 17.26\% (Energy), etc. \\
	\href{https://github.com/h-jiashu/NGLS-Diff}{\textcolor{blue}{NGLS-Diff}} \shortcite{han2024diffusion} & C & SWaT, MSL, etc. & F1: 97.52\% (SWaT), 90.81\% (MSL), 95.30\% (SMAP), 98.04\% (PSM) \\
	\href{https://github.com/ForestsKing/D3R}{\textcolor{blue}{$D^{3}R$}} \shortcite{wang2023drift} & C & PSM, SMD, SWaT & F1: 76.09\% (PSM), 86.82\% (SMD), 78.12\% (SWaT) \\
	\href{https://github.com/ChunjingXiao/DiffAD}{\textcolor{blue}{DiffAD}} \shortcite{xiao2023imputationbased} & C & MSL, SWaT, etc. & F1: 94.19\% (MSL), 97.66\% (SWaT), 97.95\% (PSM), 96.95\% (SMAP), etc. \\
	DiffADT \shortcite{zuo2024unsupervised} & C & SMAP, MSL, SMD & Pre: 96.33\% (SMAP), 95.21\% (MSL), 79.87\% (SMD), etc.  \\ \bottomrule
	\end{tabular}}}
    \caption{Summary of TSAD methods with metric results on datasets.}
	\label{tab:2}
    \vspace{-0.3cm}
	\end{table}	

DMs have emerged as a powerful technique for TSAD, effectively learning complex distributions of sequential data \cite{yang2024survey}. DMs can identify various anomaly types, including point, contextual, and collective anomalies \cite{sui2024anomaly}. Advanced approaches like NGLS-Diff \cite{han2024diffusion} leverage latent spaces, while methods \cite{zuo2024unsupervised} used structured state space layers explicitly to model temporal dependencies. Recently, some innovations address key challenges: D3R \cite{wang2023drift} tackles non-stationarity through dynamic decomposition, while TimeDiT \cite{cao2024timedit} introduces a foundation model approach with multiple masking schemes. Another key challenge is non-stationarity. DiffAD \cite{xiao2023imputationbased} handle missing data through imputation, complemented by self-supervised techniques \cite{liu2024selfsupervision} and co-evolving strategies \cite{lee2023codi} for mixed-type temporal data.

\subsection{Video Anomaly Detection}\label{sec4.3}
\begin{table}[tb!]
	\centering
	\setlength{\tabcolsep}{1mm}{
	\resizebox{0.48\textwidth}{!}{
		\begin{tabular}{@{}>{\centering\arraybackslash}p{2.6cm}|c|>{\centering\arraybackslash}p{3.1cm}|m{5.2cm}@{}}

		\toprule
		\multicolumn{1}{c|}{\textbf{Method}} & \multicolumn{1}{c|}{\textbf{DS}} & \multicolumn{1}{c|}{\textbf{Datasets}} & \multicolumn{1}{c}{\textbf{Results of metrics on datasets}}\\ \midrule
	\href{https://github.com/aleflabo/MoCoDAD}{\textcolor{blue}{MoCoDAD}} \shortcite{flaborea2023multimodal} & D & UB, HR-STC, etc. & AUC: 77.6\% (HR-STC), 89.0\% (HR-Ave), 68.4\% (HR-UB), 68.3\% (UB) \\
	\cite{tur2023unsupervised} & D & UCF, ShT & AUC: 76.36\% (ShT), 63.67\% (UCF) \\
	DHVAD \shortcite{cheng2024denoising} & C & Ped2, Ave, ShT & AUC: 98.1\% (Ped2), 88.3\% (Ave), 78.9\% (ShT) \\
	\href{https://github.com/yinyjin/DualAnoDiff}{\textcolor{blue}{FDAE}} \shortcite{zhu2024flowguided} & D & Ave, ShT, Ped2 & AUC: 97.7\% (Ped2), 91.1\% (Ave), 73.8\% (ShT) \\
	\href{https://github.com/aleflabo/MoCoDAD}{\textcolor{blue}{Masked Diffusion}} \shortcite{fang2023masked} & D & CrossTask,NIV,COIN & SR: 39.17\% (CrossTask), 32.35\% (NIV), 29.43\% (COIN), etc. \\
	FPDM \shortcite{yan2023feature} & D & Ave, ShT, UCF, UB & AUC: 90.1\% (Ave), 78.6\% (ShT), 74.7\% (UCF), 62.7\% (UB) \\
	\cite{wang2023ensemble} & D & Ped2, Ave, ShT & AUC: 96.5\% (Ped2), 92.2\% (Ave), 75.4\% (ShT) \\
	DiffVAD \shortcite{zhang2024safeguarding} & D & Ave, ShT, UCF, etc. & AUC: 81.9\% (ShT), 90.3\% (Ave), 87.6\% (Ped2), etc. \\
	\href{https://github.com/ffzzy840304/Masked-PDPP}{\textcolor{blue}{Masked Diffusion}} \shortcite{fang2023masked} & C & CrossTask,NIV,COIN & SR: 39.17\% (CrossTask), 23.47\% (CrossTask), 32.35\% (NIV), etc. \\ 
	\href{https://github.com/LHaoooo/VADiffusion}{\textcolor{blue}{VADiffusion}} \shortcite{liu2024vadiffusion} & D & Ped2, Ave, ShT & AUC: 98.2\% (Ped2), 94.93\% (Ave), 97.32\% (ShT) \\ \bottomrule
	\end{tabular}}}
    \caption{Summary of VAD methods with metric results on datasets.}
	\label{tab:3}
	\end{table}

VAD effectively leverages spatio-temporal information to identify unusual events. DMs have emerged as a highly promising VAD approach due to their ability to learn complex data distributions and generate high-quality reconstructions. For example, \cite{tur2023exploring} investigated unsupervised VAD using DMs, relying on high reconstruction error to indicate anomalies, while \cite{tur2023unsupervised} enhanced this approach by introducing compact motion representations as conditional information. Several innovative architectures have been recently proposed to improve detection accuracy: VADiffusion \cite{liu2024vadiffusion} employs a dual-branch structure combining motion vector reconstruction and I-frame prediction, while FDAE \cite{zhu2024flowguided} introduces a flow-guided diffusion autoencoder with sample refinement for comprehensive detection of both appearance and motion anomalies. Recent advances include \cite{yan2023feature}'s feature prediction diffusion model and \cite{cheng2024denoising}'s denoising diffusion-augmented hybrid framework, both enhancing semantic understanding of normal patterns. Furthermore, \cite{wang2023ensemble} proposed an ensemble approach using stochastic reconstructions and motion filters, while \cite{fang2023masked} explored masked diffusion with task-awareness for more focused anomaly detection in specific contexts.

\subsection{Multimodal Anomaly Detection}\label{sec4.4}
\begin{table}[tb!]
    \centering
    \setlength{\tabcolsep}{0.5mm}{
    \resizebox{0.48\textwidth}{!}{
    \begin{tabular}{@{}>{\centering\arraybackslash}p{2.5cm}|>{\centering\arraybackslash}p{1cm}|c|>{\centering\arraybackslash}p{3cm}|m{5.8cm}@{}}
    \toprule
    \multicolumn{1}{c|}{\textbf{Method}} & \multicolumn{1}{c|}{\textbf{Mods.}} & \multicolumn{1}{c|}{\textbf{DS}} & \multicolumn{1}{c|}{\textbf{Datasets}} & \multicolumn{1}{c}{\textbf{Results of metrics on datasets}} \\ \midrule
    \href{https://github.com/swyoon/manifold-projection-diffusion-recovery-pytorch}{\textcolor{blue}{MPDR}} \shortcite{yoon2023energybased} & I, V, A & C &MNIST, CIFAR-10, etc. &AUPR: 76.40\% (MNIST), 98.60\% (CIFAR-10), 83.38\% (CIFAR-100), etc. \\
    \href{https://github.com/yinyjin/DualAnoDiff}{\textcolor{blue}{DIAG}} \shortcite{capogrosso2024exploiting} & I, T & C & KSDD2 & AP: 80.1\% (zero-shot), 92.4\% (full-shot) \\
    \href{https://github.com/hujiecpp/MVTec-Caption}{\textcolor{blue}{AnomalyXFusion}} \shortcite{hu2024anomalyxfusion} & I, T, Te & C & MVTec-AD, MVTec LOCO & IS: 1.82 (MVTec-AD), IC-LPIPS: 0.33 (MVTec-AD), etc \\ 
    \bottomrule
    \end{tabular}}}
    \fontsize{8.5pt}{8pt}\selectfont {
    \begin{minipage}{0.48\textwidth}
    \textit{Notes:} \textbf{Mods.} indicates modalities, where 'I' is image, 'V' is vector, 'A' is audio, 'T' is text, and 'Te' is texture.
    \end{minipage}
    }
    \caption{Summary of MAD methods with metric results on datasets.}
    \label{tab:4}
    \vspace{-0.3cm}
    \end{table}

MAD integrates data from various sources to identify deviations from expected behavior. While DMs are relatively nascent in this area, they hold substantial promise through advanced multimodal data integration. For instance, AnomalyXFusion \cite{hu2024anomalyxfusion} enhances anomaly synthesis by effectively combining image, text, and mask features, while \cite{flaborea2023multimodal} demonstrates significant success in skeleton-based VAD through motion-conditioned diffusion models. For IAD, combining visual data with text descriptions significantly improves subtle anomaly detection \cite{capogrosso2024exploiting}, with energy-based approaches \cite{yoon2023energybased} leveraging manifold structures for more accurate boundary learning. GLAD \cite{yao2025glad} further suggests extending DMs for multimodal data through specialized architectures, while integration with LLMs offers promising directions for context-aware anomaly detection.
\section{Discussions}\label{sec5}
Despite significant progress in DMAD, several critical challenges remain. In this section, we explore key areas that require further research and development.
\paragraph{Computational Cost.}\label{sec5.1}
Widespread adoption of DMAD remains constrained by substantial computational requirements, particularly when processing high-dimensional data or extended time series. Computational demands manifest during both distribution learning in training and data generation in sampling phases, with the sampling process's iterative nature demanding numerous steps for quality output \cite{cui2023elucidating}. For example, high-resolution IAD suffers from increased computational overhead due to data volume, model complexity, and the need to capture temporal dependencies \cite{wyatt2022anoddpm,sui2024anomaly}.  However, some researchers are actively exploring solutions.  One promising direction is faster sampling methods, such as progressive distillation and optimized sampling schedules like align your steps, which aim to reduce sampling steps while maintaining quality \cite{salimans2021progressive}.  Another approach is model compression through techniques like pruning and quantization.  Additionally, efficient architectures, like \cite{cui2023elucidating} leveraging ER SDEs for faster sampling, and preconditioning methods offer further computational gains.

\paragraph{Interpretability and Explainability.}\label{sec5.2}
Interpretability remains a key challenge for DMAD \cite{katsuoka2024statistical}.  Understanding DM's decision-making is crucial, especially in critical applications~\cite{navalmarimont2024ensembled}.  Visualizing anomaly scores, like spatially highlighting anomalous image regions as in reconstruction-based methods \cite{yao2025glad}, becomes essential.  However, the iterative denoising process inherent in DMs complicates interpretability.  Explaining anomaly detection requires identifying and explaining deviating features.  Integrating DMs with explainable AI (xAI) techniques, such as incorporating attention mechanisms or leveraging LLMs for textual/visual explanations \cite{wang2023odd}, offers a promising research direction.

\paragraph{Complex Data Distributions.}\label{sec5.3}
A key challenge in applying DMAD lies in handling complex data distributions.  Imbalanced datasets, where anomalies are rare, can bias DMs towards the majority class \cite{yang2024novel}, hindering accurate anomaly modeling.  Similarly, multimodal datasets, representing distinct normal behaviors, can confound DM learning \cite{zuo2024unsupervised}, potentially misclassifying data from less prominent modes. Noisy or missing data further complicates DM training and inference \cite{choi2022perception}, as differentiating true anomalies from data imperfections becomes difficult.  However, potential solutions exist. Data augmentation techniques can address class imbalance by generating synthetic minority class samples \cite{yang2024novel}. Robust training methods, such as using semi-unbalanced optimal transport, can enhance DM resilience to noise and outliers \cite{dao2024highquality}. Specialized architectures, potentially incorporating mechanisms for handling missing data or modeling multiple modes \cite{xiao2023imputationbased}, and dynamic step size computation \cite{tebbe2024dynamic} could further improve DM performance on complex distributions.

\paragraph{Robustness and Adversarial Attacks.}\label{sec5.4}
Adversarial robustness is a critical concern for DMAD.  Similar to other deep learning models, DMs are vulnerable to adversarial perturbations, raising concerns about the reliability of diffusion-based AD systems \cite{chen2024diffilter}.  For example,  \cite{kang2025diffender} demonstrates the potential of DMs for adversarial defense in classification, directly applying such methods to AD can reduce anomaly detection rates. Specifically, the purification process may remove crucial anomaly signals along with noise.  Additionally, adversarial examples exhibit misalignment within DM manifolds, offering a potential detection avenue but requiring further investigation.  The observed robustness differences between pixel-space diffusion models (PDMs) and more vulnerable latent diffusion models (LDMs) \cite{graham2023unsupervised} further underscore the need to consider specific DM types.  Consequently, future research should prioritize robust DM-based AD methods, including architectures and training procedures that distinguish genuine anomalies from adversarial noise.  Some promising directions include defense strategies like DIFFender \cite{kang2025diffender}, leveraging text-guided diffusion and the adversarial anomaly perception phenomenon, and incorporating insights from adversarial example behavior within DM manifolds.

\paragraph{Edge-Cloud Collaboration.}\label{sec5.5}
Real-time DMAD faces significant adoption barriers due to their substantial computational demands \cite{li2023fast}. Edge-cloud collaboration offers a promising solution to this challenge by distributing the workload between edge devices and cloud servers. Carefully designed Lightweight DMs deployed on edge devices perform initial anomaly screening, while resource-intensive tasks like full reconstructions are efficiently offloaded to cloud servers \cite{yan2024hybrid}. In addition, federated learning enables collaborative model training across edge devices and the cloud without sharing sensitive data, thereby enhancing generalization and preserving privacy~\cite{jin2024survey}. Dynamic DM partitioning \cite{chen2024dynamic} facilitates adaptive resource allocation, optimizing performance based on network conditions and computational demands. Strategic data placement and distributed DNN deployment principles further enhance system efficiency. The integration of proactive detection mechanisms, exemplified by Maat \cite{lee2023maat}, proves particularly valuable for time-critical applications in cloud monitoring and AIOps environments.  

\paragraph{Integrating with Large Language Models.}\label{sec5.6}
Integrating DMs with LLMs offers a highly promising avenue for enhancing anomaly detection, particularly by generating human-interpretable explanations of detected anomalies and incorporating rich contextual information \cite{kumar2023selfsupervised}. For example, LLMs can leverage detailed textual descriptions to provide context for observed fluctuations, distinguishing genuine anomalies from expected variations for MAD \cite{capogrosso2024exploiting}. However, current LLM integration for anomaly detection faces a key challenge: effectively representing and tokenizing temporal data. Existing tokenizers, primarily designed for text, may not adequately capture the subtle nuances of numerical and temporal data, potentially hindering performance \cite{li2024selfsupervised}. Consequently, a critical research direction involves developing efficient data representation techniques and specialized LLMs for anomaly detection to fully realize this synergistic approach's potential \cite{tebbe2024dynamic}.
\vspace{-0.3cm}
\section{Conclusion}\label{sec6}
In this survey, we first introduce the core concepts of anomaly detection and diffusion models, providing a foundation for understanding DMAD. Then, we systematically review existing methodologies and their tasks across diverse data types, including image, time series, and multimodal data. Furthermore, we analyze representative approaches for each data type, highlighting their strengths and limitations. To promote further research, we identify key challenges and promising future directions, aiming to advance DMAD research and inspire innovative solutions for real-world applications.
\bibliographystyle{named}
\begin{spacing}{0.83} 
\bibliography{00_ijcai25}

\begin{thebibliography}{}

\bibitem[\protect\citeauthoryear{Bercea \bgroup \em et al.\egroup }{2023}]{bercea2023mask}
Cosmin~I. Bercea, Michael Neumayr, Daniel Rueckert, and Julia~A. Schnabel.
\newblock Mask, stitch, and re-sample: Enhancing robustness and generalizability in anomaly detection through automatic diffusion models.
\newblock {\em arXiv:2305.19643}, 2023.

\bibitem[\protect\citeauthoryear{Cao \bgroup \em et al.\egroup }{2024}]{cao2024timedit}
Defu Cao, Wen Ye, and Yan Liu.
\newblock Timedit: General-purpose diffusion transformers for time series foundation model.
\newblock In {\em ICMLW}, 2024.

\bibitem[\protect\citeauthoryear{Capogrosso \bgroup \em et al.\egroup }{2024}]{capogrosso2024exploiting}
Luigi Capogrosso, Alvise Vivenza, Andrea Chiarini, Francesco Setti, and Marco Cristani.
\newblock Exploiting multimodal latent diffusion models for accurate anomaly detection in industry 5.0.
\newblock In {\em xAI}, volume 3762, pages 230--235, 2024.

\bibitem[\protect\citeauthoryear{Chen \bgroup \em et al.\egroup }{2024a}]{chen2024dynamic}
Xiangchun Chen, Jiannong Cao, Yuvraj Sahni, Shan Jiang, and Zhixuan Liang.
\newblock Dynamic task offloading in edge computing based on dependency-aware reinforcement learning.
\newblock {\em IEEE TCC}, 12:594--608, 2024.

\bibitem[\protect\citeauthoryear{Chen \bgroup \em et al.\egroup }{2024b}]{chen2024diffilter}
Yong Chen, Xuedong Li, Peng Hu, Dezhong Peng, and Xu~Wang.
\newblock Diffilter: Defending against adversarial perturbations with diffusion filter.
\newblock {\em IEEE TIFS}, 19:6779--6794, 2024.

\bibitem[\protect\citeauthoryear{Cheng \bgroup \em et al.\egroup }{2024}]{cheng2024denoising}
Kai Cheng, Yaning Pan, Yang Liu, Xinhua Zeng, and Rui Feng.
\newblock Denoising diffusion-augmented hybrid video anomaly detection via reconstructing noised frames.
\newblock In {\em IJCAI}, volume~2, pages 695--703, 2024.

\bibitem[\protect\citeauthoryear{Choi \bgroup \em et al.\egroup }{2022}]{choi2022perception}
Jooyoung Choi, Jungbeom Lee, Chaehun Shin, Sungwon Kim, Hyunwoo Kim, and Sungroh Yoon.
\newblock Perception prioritized training of diffusion models.
\newblock In {\em CVPR}, pages 11472--11481, 2022.

\bibitem[\protect\citeauthoryear{Cui \bgroup \em et al.\egroup }{2023}]{cui2023elucidating}
Qinpeng Cui, Xinyi Zhang, Zongqing Lu, and Qingmin Liao.
\newblock Elucidating the solution space of extended reverse-time sde for diffusion models, 2023.

\bibitem[\protect\citeauthoryear{Dao \bgroup \em et al.\egroup }{2024}]{dao2024highquality}
Quan Dao, Binh Ta, Tung Pham, and Anh Tran.
\newblock A high-quality robust diffusion framework for corrupted dataset, 2024.

\bibitem[\protect\citeauthoryear{Deveney \bgroup \em et al.\egroup }{2023}]{deveney2023closing}
Teo Deveney, Jan Stanczuk, Lisa~Maria Kreusser, Chris Budd, and Carola-Bibiane Sch{\"o}nlieb.
\newblock Closing the ode-sde gap in score-based diffusion models through the fokker-planck equation.
\newblock {\em arXiv:2311.15996}, 2023.

\bibitem[\protect\citeauthoryear{Fang \bgroup \em et al.\egroup }{2023}]{fang2023masked}
Fen Fang, Yun Liu, Ali Koksal, Qianli Xu, and Joo-Hwee Lim.
\newblock Masked diffusion with task-awareness for procedure planning in instructional videos.
\newblock {\em arXiv:2309.07409}, 2023.

\bibitem[\protect\citeauthoryear{Flaborea \bgroup \em et al.\egroup }{2023}]{flaborea2023multimodal}
Alessandro Flaborea, Luca Collorone, Guido~Maria D'Amely Di~Melendugno, Stefano D'Arrigo, Bardh Prenkaj, and Fabio Galasso.
\newblock Multimodal motion conditioned diffusion model for skeleton-based video anomaly detection.
\newblock In {\em ICCV}, pages 1--9, 2023.

\bibitem[\protect\citeauthoryear{Fontanella \bgroup \em et al.\egroup }{2024}]{fontanella2024diffusion}
Alessandro Fontanella, Grant Mair, Joanna Wardlaw, Emanuele Trucco, and Amos Storkey.
\newblock Diffusion models for counterfactual generation and anomaly detection in brain images.
\newblock {\em IEEE TMC}, 2024.

\bibitem[\protect\citeauthoryear{Graham \bgroup \em et al.\egroup }{2023}]{graham2023unsupervised}
Mark~S. Graham, Walter Hugo~Lopez Pinaya, Paul Wright, Petru-Daniel Tudosiu, Yee~H. Mah, James~T. Teo, H.~Rolf J{\"a}ger, David Werring, Parashkev Nachev, Sebastien Ourselin, and M.~Jorge Cardoso.
\newblock Unsupervised 3d out-of-distribution detection with~latent diffusion models.
\newblock In {\em MICCAI}, pages 446--456, 2023.

\bibitem[\protect\citeauthoryear{Han \bgroup \em et al.\egroup }{2024a}]{han2024diffusion}
Jiashu Han, Shanshan Feng, Min Zhou, Xinyu Zhang, Yew~Soon Ong, and Xutao Li.
\newblock Diffusion model in normal gathering latent space for time series anomaly detection.
\newblock In {\em ECML PKDD}, volume 14943, pages 284--300, 2024.

\bibitem[\protect\citeauthoryear{Han \bgroup \em et al.\egroup }{2024b}]{han2024neural}
Yinbin Han, Meisam Razaviyayn, and Renyuan Xu.
\newblock Neural network-based score estimation in diffusion models: Optimization and generalization, 2024.

\bibitem[\protect\citeauthoryear{Hu and Jin}{2023}]{hu2023anodode}
Xianyao Hu and Congming Jin.
\newblock Anodode: Anomaly detection with diffusion ode, 2023.

\bibitem[\protect\citeauthoryear{Hu \bgroup \em et al.\egroup }{2024a}]{hu2024anomalyxfusion}
Jie Hu, Yawen Huang, Yilin Lu, Guoyang Xie, Guannan Jiang, Yefeng Zheng, and Zhichao Lu.
\newblock Anomalyxfusion: Multi-modal anomaly synthesis with diffusion.
\newblock {\em arXiv:2404.19444}, 2024.

\bibitem[\protect\citeauthoryear{Hu \bgroup \em et al.\egroup }{2024b}]{hu2024anomalydiffusion}
Teng Hu, Jiangning Zhang, Ran Yi, Yuzhen Du, Xu~Chen, Liang Liu, Yabiao Wang, and Chengjie Wang.
\newblock Anomalydiffusion: Few-shot anomaly image generation with diffusion model.
\newblock In {\em AAAI}, volume~38, pages 8526--8534, 2024.

\bibitem[\protect\citeauthoryear{Iqbal \bgroup \em et al.\egroup }{2024}]{iqbal2024unsupervised}
Hasan Iqbal, Umar Khalid, Chen Chen, and Jing Hua.
\newblock Unsupervised anomaly detection in~medical images using masked diffusion model.
\newblock In {\em MLMI}, pages 372--381, 2024.

\bibitem[\protect\citeauthoryear{Jin \bgroup \em et al.\egroup }{2024}]{jin2024survey}
Ming Jin, Huan~Yee Koh, Qingsong Wen, Daniele Zambon, Cesare Alippi, Geoffrey~I. Webb, Irwin King, and Shirui Pan.
\newblock A survey on graph neural networks for time series: Forecasting, classification, imputation, and anomaly detection.
\newblock {\em IEEE TPAMI}, 46(12):1--20, 2024.

\bibitem[\protect\citeauthoryear{Kang \bgroup \em et al.\egroup }{2025}]{kang2025diffender}
Caixin Kang, Yinpeng Dong, Zhengyi Wang, Shouwei Ruan, Yubo Chen, Hang Su, and Xingxing Wei.
\newblock Diffender: Diffusion-based adversarial defense against patch attacks.
\newblock In {\em ECCV}, pages 130--147, 2025.

\bibitem[\protect\citeauthoryear{Katsuoka \bgroup \em et al.\egroup }{2024}]{katsuoka2024statistical}
Teruyuki Katsuoka, Tomohiro Shiraishi, Daiki Miwa, Vo~Nguyen~Le Duy, and Ichiro Takeuchi.
\newblock Statistical test on diffusion model-based anomaly detection by selective inference, 2024.

\bibitem[\protect\citeauthoryear{Kumar \bgroup \em et al.\egroup }{2023}]{kumar2023selfsupervised}
Komal Kumar, Snehashis Chakraborty, and Sudipta Roy.
\newblock Self-supervised diffusion model for anomaly segmentation in medical imaging.
\newblock In {\em PRMI}, volume 14301, pages 359--368, 2023.

\bibitem[\protect\citeauthoryear{Le~Lan and Dinh}{2021}]{lelan2021perfect}
Charline Le~Lan and Laurent Dinh.
\newblock Perfect density models cannot guarantee anomaly detection.
\newblock {\em Entropy}, 23:1690, 2021.

\bibitem[\protect\citeauthoryear{Lee \bgroup \em et al.\egroup }{2023a}]{lee2023codi}
Chaejeong Lee, Jayoung Kim, and Noseong Park.
\newblock Codi: Co-evolving contrastive diffusion models for mixed-type tabular synthesis.
\newblock In {\em ICML}, pages 18940--18956, 2023.

\bibitem[\protect\citeauthoryear{Lee \bgroup \em et al.\egroup }{2023b}]{lee2023maat}
Cheryl Lee, Tianyi Yang, Zhuangbin Chen, Yuxin Su, and Michael~R. Lyu.
\newblock Maat: Performance metric anomaly anticipation for cloud services with conditional diffusion.
\newblock In {\em ASE}, pages 116--128, 2023.

\bibitem[\protect\citeauthoryear{Li \bgroup \em et al.\egroup }{2023}]{li2023fast}
Jinpeng Li, Hanqun Cao, Jiaze Wang, Furui Liu, Qi~Dou, Guangyong Chen, and Pheng-Ann Heng.
\newblock Fast non-markovian diffusion model for~weakly supervised anomaly detection in~brain mr images.
\newblock In {\em MICCAI}, volume 14224, pages 579--589, 2023.

\bibitem[\protect\citeauthoryear{Li \bgroup \em et al.\egroup }{2024}]{li2024selfsupervised}
Shu Li, Jiong Yu, Yi~Lu, Guangqi Yang, Xusheng Du, and Su~Liu.
\newblock Self-supervised enhanced denoising diffusion for anomaly detection.
\newblock {\em Inf. Sci.}, 669:120612, 2024.

\bibitem[\protect\citeauthoryear{Liu \bgroup \em et al.\egroup }{2024a}]{liu2024vadiffusion}
Hao Liu, Lijun He, Miao Zhang, and Fan Li.
\newblock Vadiffusion: Compressed domain information guided conditional diffusion for video anomaly detection.
\newblock {\em IEEE TCSVT}, 34:8398--8411, 2024.

\bibitem[\protect\citeauthoryear{Liu \bgroup \em et al.\egroup }{2024b}]{liu2024selfsupervision}
Yixin Liu, Thalaiyasingam Ajanthan, Hisham Husain, and Vu~Nguyen.
\newblock Self-supervision improves diffusion models for tabular data imputation.
\newblock In {\em CIKM}, pages 1513--1522, 2024.

\bibitem[\protect\citeauthoryear{Livernoche \bgroup \em et al.\egroup }{2023}]{livernoche2023diffusion}
Victor Livernoche, Vineet Jain, Yashar Hezaveh, and Siamak Ravanbakhsh.
\newblock On diffusion modeling for anomaly detection.
\newblock In {\em ICLR}, 2023.

\bibitem[\protect\citeauthoryear{Luo}{2023}]{luo2023comprehensive}
Weijian Luo.
\newblock A comprehensive survey on knowledge distillation of diffusion models, 2023.

\bibitem[\protect\citeauthoryear{Naval~Marimont \bgroup \em et al.\egroup }{2024}]{navalmarimont2024ensembled}
Sergio Naval~Marimont, Vasilis Siomos, Matthew Baugh, Christos Tzelepis, Bernhard Kainz, and Giacomo Tarroni.
\newblock Ensembled cold-diffusion restorations for~unsupervised anomaly detection.
\newblock In {\em MICCAI}, pages 243--253, 2024.

\bibitem[\protect\citeauthoryear{Pang \bgroup \em et al.\egroup }{2021}]{pang2021deepa}
Guansong Pang, Chunhua Shen, Longbing Cao, and Anton Van~Den Hengel.
\newblock Deep learning for anomaly detection: A review.
\newblock {\em ACM CSUR}, 54(2):38:1--38:38, March 2021.

\bibitem[\protect\citeauthoryear{Salimans and Ho}{2021}]{salimans2021progressive}
Tim Salimans and Jonathan Ho.
\newblock Progressive distillation for fast sampling of diffusion models.
\newblock In {\em ICLR}, 2021.

\bibitem[\protect\citeauthoryear{Sheng \bgroup \em et al.\egroup }{2024}]{sheng2024surface}
Xinyu Sheng, Shande Tuo, and Lu~Wang.
\newblock Surface anomaly detection and localization with diffusion-based reconstruction.
\newblock In {\em IJCNN}, pages 1--8, 2024.

\bibitem[\protect\citeauthoryear{Sui \bgroup \em et al.\egroup }{2024}]{sui2024anomaly}
Jialin Sui, Jinsong Yu, Yue Song, and Jian Zhang.
\newblock Anomaly detection for telemetry time series using a denoising diffusion probabilistic model.
\newblock {\em IEEE Sens. J.}, 24:16429--16439, 2024.

\bibitem[\protect\citeauthoryear{Tebbe and Tayyub}{2024}]{tebbe2024dynamic}
Justin Tebbe and Jawad Tayyub.
\newblock Dynamic addition of noise in a diffusion model for anomaly detection.
\newblock In {\em CVPR}, pages 3940--3949, 2024.

\bibitem[\protect\citeauthoryear{Tur \bgroup \em et al.\egroup }{2023a}]{tur2023exploring}
Anil~Osman Tur, Nicola Dall'Asen, Cigdem Beyan, and Elisa Ricci.
\newblock Exploring diffusion models for unsupervised video anomaly detection.
\newblock In {\em ICIP}, pages 2540--2544, 2023.

\bibitem[\protect\citeauthoryear{Tur \bgroup \em et al.\egroup }{2023b}]{tur2023unsupervised}
Anil~Osman Tur, Nicola Dall'Asen, Cigdem Beyan, and Elisa Ricci.
\newblock Unsupervised video anomaly detection with~diffusion models conditioned on~compact motion representations.
\newblock In {\em ICIAP}, pages 49--62, 2023.

\bibitem[\protect\citeauthoryear{Wang and Vastola}{2023}]{wang2023hidden}
Binxu Wang and John~J. Vastola.
\newblock The hidden linear structure in score-based models and its application.
\newblock {\em arXiv:2311.10892}, 2023.

\bibitem[\protect\citeauthoryear{Wang \bgroup \em et al.\egroup }{2023a}]{wang2023drift}
Chengsen Wang, Zirui Zhuang, Qi~Qi, Jingyu Wang, Xingyu Wang, Haifeng Sun, and Jianxin Liao.
\newblock Drift doesn't matter: Dynamic decomposition with diffusion reconstruction for unstable multivariate time series anomaly detection.
\newblock In {\em NeurIPS}, volume~36, pages 10758--10774, 2023.

\bibitem[\protect\citeauthoryear{Wang \bgroup \em et al.\egroup }{2023b}]{wang2023odd}
He~Wang, Longquan Dai, Jinglin Tong, and Yan Zhai.
\newblock Odd: One-class anomaly detection via the diffusion model.
\newblock In {\em ICIP}, pages 3000--3004, 2023.

\bibitem[\protect\citeauthoryear{Wang \bgroup \em et al.\egroup }{2023c}]{wang2023diffusion}
Wenjie Wang, Yiyan Xu, Fuli Feng, Xinyu Lin, Xiangnan He, and Tat-Seng Chua.
\newblock Diffusion recommender model.
\newblock In {\em SIGIR}, pages 832--841, 2023.

\bibitem[\protect\citeauthoryear{Wang \bgroup \em et al.\egroup }{2023d}]{wang2023ensemble}
Zhiqiang Wang, Xiaojing Gu, Jingyu Hu, and Xingsheng Gu.
\newblock Ensemble anomaly score for video anomaly detection using denoise diffusion model and motion filters.
\newblock {\em Neurocomputing}, 553:126589, 2023.

\bibitem[\protect\citeauthoryear{Wu \bgroup \em et al.\egroup }{2024}]{wu2024unsupervised}
Di~Wu, Shicai Fan, Xue Zhou, Li~Yu, Yuzhong Deng, Jianxiao Zou, and Baihong Lin.
\newblock Unsupervised anomaly detection via masked diffusion posterior sampling.
\newblock {\em arXiv:2404.17900}, 2024.

\bibitem[\protect\citeauthoryear{Wyatt \bgroup \em et al.\egroup }{2022}]{wyatt2022anoddpm}
Julian Wyatt, Adam Leach, Sebastian~M. Schmon, and Chris~G. Willcocks.
\newblock Anoddpm: Anomaly detection with denoising diffusion probabilistic models using simplex noise.
\newblock In {\em CVPR}, pages 650--656, 2022.

\bibitem[\protect\citeauthoryear{Xiao \bgroup \em et al.\egroup }{2023}]{xiao2023imputationbased}
Chunjing Xiao, Zehua Gou, Wenxin Tai, Kunpeng Zhang, and Fan Zhou.
\newblock Imputation-based time-series anomaly detection with conditional weight-incremental diffusion models.
\newblock In {\em KDD}, pages 1--9, 2023.

\bibitem[\protect\citeauthoryear{Xu \bgroup \em et al.\egroup }{2023}]{xu2023unsupervised}
Haohao Xu, Shuchang Xu, and Wenzhen Yang.
\newblock Unsupervised industrial anomaly detection with diffusion models.
\newblock {\em J. Vis. Commun. Image Represent.}, 97:103983, 2023.

\bibitem[\protect\citeauthoryear{Yan \bgroup \em et al.\egroup }{2023}]{yan2023feature}
Cheng Yan, Shiyu Zhang, Yang Liu, Guansong Pang, and Wenjun Wang.
\newblock Feature prediction diffusion model for video anomaly detection.
\newblock In {\em ICCV}, pages 5504--5514, 2023.

\bibitem[\protect\citeauthoryear{Yan \bgroup \em et al.\egroup }{2024}]{yan2024hybrid}
Chenqian Yan, Songwei Liu, Hongjian Liu, Xurui Peng, Xiaojian Wang, Fangmin Chen, Lean Fu, and Xing Mei.
\newblock Hybrid sd: Edge-cloud collaborative inference for stable diffusion models.
\newblock {\em arXiv:2408.06646}, 2024.

\bibitem[\protect\citeauthoryear{Yang \bgroup \em et al.\egroup }{2024a}]{yang2024novel}
Xiongyan Yang, Tianyi Ye, Xianfeng Yuan, Weijie Zhu, Xiaoxue Mei, and Fengyu Zhou.
\newblock A novel data augmentation method based on denoising diffusion probabilistic model for fault diagnosis under imbalanced data.
\newblock {\em IEEE TII}, 20:7820--7831, 2024.

\bibitem[\protect\citeauthoryear{Yang \bgroup \em et al.\egroup }{2024b}]{yang2024survey}
Yiyuan Yang, Ming Jin, Haomin Wen, Chaoli Zhang, Yuxuan Liang, Lintao Ma, Yi~Wang, Chenghao Liu, Bin Yang, Zenglin Xu, Jiang Bian, Shirui Pan, and Qingsong Wen.
\newblock A survey on diffusion models for time series and spatio-temporal data, 2024.

\bibitem[\protect\citeauthoryear{Yao \bgroup \em et al.\egroup }{2025}]{yao2025glad}
Hang Yao, Ming Liu, Zhicun Yin, Zifei Yan, Xiaopeng Hong, and Wangmeng Zuo.
\newblock Glad: Towards better reconstruction with~global and~local adaptive diffusion models for~unsupervised anomaly detection.
\newblock In {\em ECCV}, pages 1--17, 2025.

\bibitem[\protect\citeauthoryear{Yoon \bgroup \em et al.\egroup }{2023}]{yoon2023energybased}
Sangwoong Yoon, Young-Uk Jin, Yung-Kyun Noh, and Frank Park.
\newblock Energy-based models for anomaly detection: A manifold diffusion recovery approach.
\newblock In {\em NeurIPS}, volume~36, pages 1--21, 2023.

\bibitem[\protect\citeauthoryear{Zhan \bgroup \em et al.\egroup }{2024}]{zhan2024enhancing}
Jiawei Zhan, Jinxiang Lai, Bin-Bin Gao, Jun Liu, Xiaochen Chen, and Chengjie Wang.
\newblock Enhancing multi-class anomaly detection via diffusion refinement with dual conditioning.
\newblock {\em arXiv:2407.01905}, 2024.

\bibitem[\protect\citeauthoryear{Zhang and Pilanci}{2024}]{zhang2024analyzing}
Fangzhao Zhang and Mert Pilanci.
\newblock Analyzing neural network-based generative diffusion models through convex optimization, 2024.

\bibitem[\protect\citeauthoryear{Zhang \bgroup \em et al.\egroup }{2023}]{zhang2023diffusionad}
Hui Zhang, Zheng Wang, Zuxuan Wu, and Yu-Gang Jiang.
\newblock Diffusionad: Norm-guided one-step denoising diffusion for anomaly detection, 2023.

\bibitem[\protect\citeauthoryear{Zhang \bgroup \em et al.\egroup }{2024}]{zhang2024safeguarding}
Menghao Zhang, Jingyu Wang, Qi~Qi, Pengfei Ren, Haifeng Sun, Zirui Zhuang, Lei Zhang, and Jianxin Liao.
\newblock Safeguarding sustainable cities: Unsupervised video anomaly detection through diffusion-based latent pattern learning.
\newblock In {\em IJCAI}, volume~8, pages 7572--7580, 2024.

\bibitem[\protect\citeauthoryear{Zhu \bgroup \em et al.\egroup }{2024}]{zhu2024flowguided}
Aoni Zhu, Wenjun Wang, and Cheng Yan.
\newblock Flow-guided diffusion autoencoder for unsupervised video anomaly detection.
\newblock In {\em PRCV}, volume 14430, pages 183--194, 2024.

\bibitem[\protect\citeauthoryear{Zuo \bgroup \em et al.\egroup }{2024}]{zuo2024unsupervised}
Haiwei Zuo, Aiqun Zhu, Yanping Zhu, Yinping Liao, Shiman Li, and Yun Chen.
\newblock Unsupervised diffusion based anomaly detection for time series.
\newblock {\em APIN}, 54:8968--8981, 2024.

\end{thebibliography}
\end{spacing}

\end{document}